\documentclass[letterpaper, 10 pt, conference]{ieeeconf}
\IEEEoverridecommandlockouts
\usepackage{cite}
\usepackage{amsmath,amssymb,amsfonts}
\usepackage{algorithmic}
\usepackage{graphicx}
\usepackage{textcomp}
\usepackage{siunitx}
\usepackage{booktabs}
\usepackage{tabularray}
\usepackage{tabulary}
\usepackage{acro}
\usepackage{subfigure} % ADDED RD 15.01.2020
\DeclareAcronym{BIM}{
  short = BIM,
  long  = building information modeling,
  short-indefinite = a,
  long-indefinite = a
}

\DeclareAcronym{MDT}{
  short = MDT,
  long  = multi-directional thrust ,
  short-indefinite = an,
  long-indefinite = a
}

\DeclareAcronym{NDT}{
  short = NDT,
  long  = nondestructive testing,
  short-indefinite = an,
  long-indefinite = a
}

\DeclareAcronym{VT}{
  short = VT,
  long  = vectored thrust
}

\DeclareAcronym{F/W}{
  short = F/W,
  long  = thrust-to-weight ratio,
  short-indefinite = an
}

\DeclareAcronym{UAV}{
  short = UAV,
  long  = unmanned aerial vehicle,
  long-indefinite = an
}

\DeclareAcronym{RC}{
  short = RC,
  long  = remote control,
  short-indefinite = an
}

\DeclareAcronym{RPM}{
  short = RPM,
  long  = revolutions per minute,
  short-indefinite = an
}

\DeclareAcronym{COM}{
  short = COM,
  long  = center of mass
}

\DeclareAcronym{VIO}{
  short = VIO,
  long  = Visual Inertial Odometry
}
\DeclareAcronym{FLU}{
  short = FLU,
  long  = Front-Left-Up
}

\DeclareAcronym{PLA}{
    short = PLA,
    long = polylactic acid filament
}

\DeclareAcronym{LiPo}{
    short =LiPo,
    long = lithium polymer battery
}
\newcommand{\reffig}[1]{Fig.~\ref{#1}}
\newcommand{\reftab}[1]{Table~\ref{#1}}
\newcommand{\refsec}[1]{Section~\ref{#1}}

\usepackage{xcolor}
\usepackage{comment}
\usepackage{stfloats}
\usepackage{mathtools} % for := (coloneqq)

\title{\LARGE \bf A perching and tilting aerial robot for precise and \\ versatile power tool work on vertical walls}
\author{Roman Dautzenberg{*}, Timo Küster{*}, Timon Mathis{*}, Yann Roth{*}, Curdin Steinauer{*} \\Gabriel Käppeli, Julian Santen, Alina Arranhado, Friederike Biffar, Till Kötter \\Christian Lanegger, Mike Allenspach, Roland Siegwart, and Rik Bähnemann
\thanks{This project was supported by Hilti AG, Armasuisse W+T, Switzerland Innovation Park Zurich, and the Swiss National Science Foundation's NCCR DFab P3.}%
\thanks{* Authors contributed equally to this work.}%
\thanks{All authors are with the Autonomous Systems Lab, ETH Zürich.}%
\thanks{Corresponding Author: R. Dautzenberg, \texttt{droman@ethz.ch}}%
}

\begin{document}
\maketitle
\thispagestyle{empty}
\pagestyle{empty}

\begin{abstract}
Drilling, grinding, and setting anchors on vertical walls are fundamental processes in everyday construction work.
Manually doing these works is error-prone, potentially dangerous, and elaborate at height.
Today, heavy mobile ground robots can perform automatic power tool work.
However, aerial vehicles could be deployed in untraversable environments and reach inaccessible places.
Existing drone designs do not provide the large forces, payload, and high precision required for using power tools.
 This work presents the first aerial robot design to perform versatile manipulation tasks on vertical concrete walls with continuous forces of up to $\mathbf{\SI{150}{\newton}}$.
The platform combines a quadrotor with active suction cups for perching on walls and a lightweight, tiltable linear tool table. This combination minimizes weight using the propulsion system for flying, surface alignment, and feed during manipulation and allows precise positioning of the power tool.
We evaluate our design in a concrete drilling application -- a challenging construction process that requires high forces, accuracy, and precision.
In $\mathbf{30}$ trials, our design can accurately pinpoint a target position despite perching imprecision.
Nine visually guided drilling experiments demonstrate a drilling precision of $\mathbf{\SI[round-mode=places,round-precision=0]{5.68}{\milli\metre}}$ without further automation.
Aside from drilling, we also demonstrate the versatility of the design by setting an anchor into concrete.
\end{abstract}
\begin{comment}
\begin{IEEEkeywords}
aerial manipulation, design, automation in construction, \acs{UAV}, robotics, perching, suction, drilling
\end{IEEEkeywords}
\end{comment}
\acresetall

\section{Introduction}
% Why automation in construction?
Construction robotics is a fast-growing market. 
This trend has multiple drivers, including workplace safety, shortage of skilled labor, material waste, construction scheduling, and higher demands in building accuracy and productivity~\cite{bogue2017prospects}.
Precise power tool work, such as drilling, is one domain where mobile robots could assist.
% Why drones?
In conjunction with \acl{BIM} and automated positioning, robots achieve millimeter accuracy referenced to the building plan~\cite{gawel2019fully}.
In particular, multi-purpose ground robots can do autonomous installations~\cite{shimizu_corporation_2018}.
However, such ground robots are limited by their reach and weight. The Shimizu Robo-Buddy, for instance, weighs \SI{1.6}{\tonne} and requires dedicated walkways to move around construction sites.
% Why drones?
On the other hand, flying robots can operate in unstructured environments and thus are an attractive alternative, especially when the payload is relatively light and inaccessible places, such as bridge pillars, must be reached.

Designing an aerial robot that does power tool work is challenging.
Four main aspects must be considered: i) handling high tool reaction forces without compromising flight stability; ii) precise and accurate tool application according to construction site standards; iii) compact mechanical construction to reduce collision risk and iv) sufficient payload to transport different power tools.
For example, the concrete drilling and anchor placement in this work requires at least \SI{80}{\newton} feed force, expects deviations of at most \SI{10}{\milli\metre}~\cite{din_18202}, and a \SI{2.3}{\kilo\gram} drill payload.
These design criteria oppose each other.
For example, adding payload reduces the \ac{F/W} and, thus, flight control response~\cite{bouabdallah2004design}.
Nevertheless, one cannot just increase the propulsion system size.
Large multi-rotors are less maneuverable~\cite{edmonds2020unmanned} and thus inconvenient on cluttered construction sites. 
Existing work typically captures one design aspect, e.g., high-force tolerant perching~\cite{kessens2019toward}, precise contact~\cite{watson2022techniques}, or compact hardware for milling~\cite{pantic2019milling}.
However, a system that combines all capabilities is still missing. 

\begin{figure}
\centerline{\includegraphics[width=0.45\textwidth]{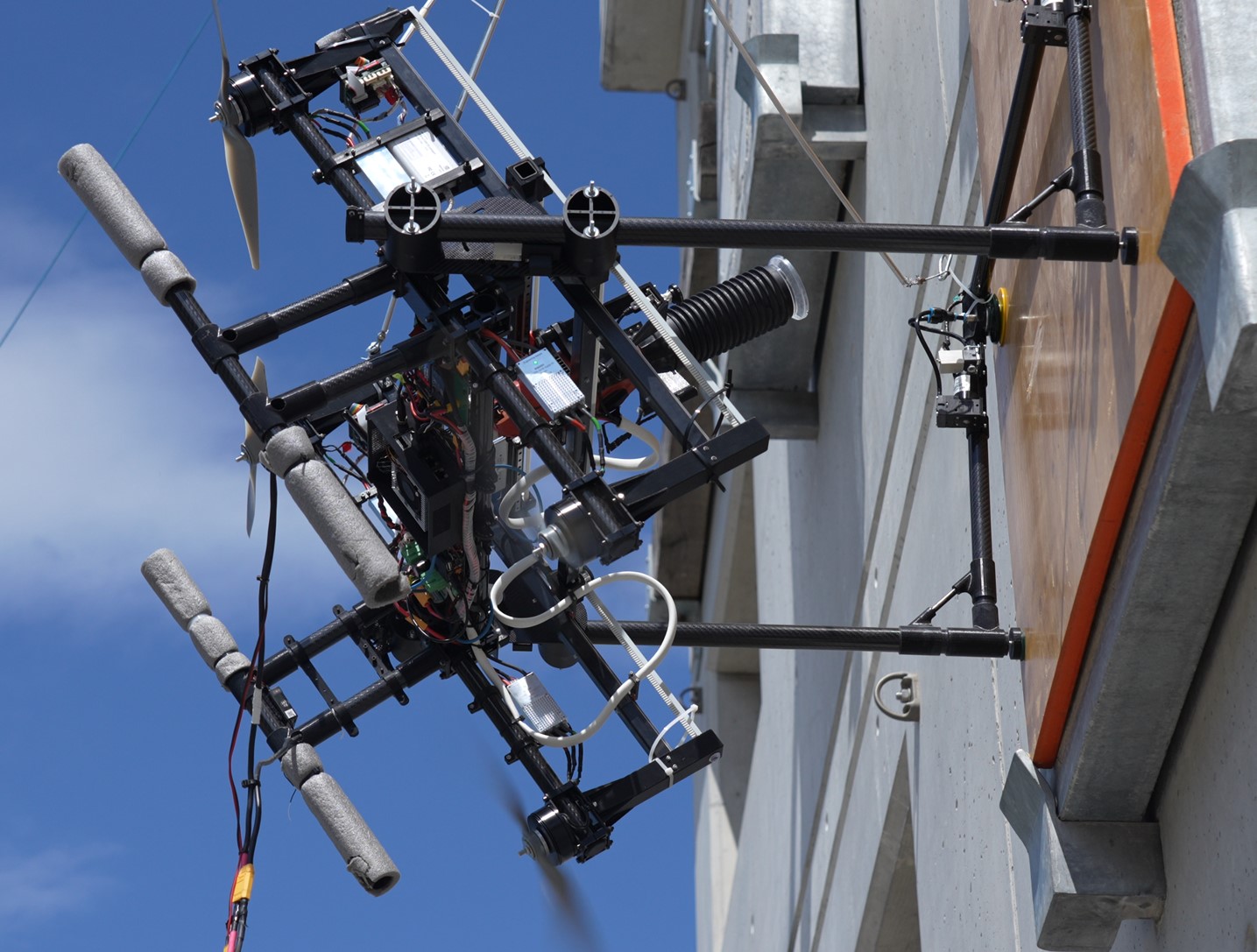}}
%\centerline{\includegraphics[width=0.45\textwidth]{aithon_teaser.jpg}}
\caption{The aerial robot perches on a wooden facade on the second floor of a building, using its front propellers to rotate its tool table for drilling.}
\label{fig:arche}
\end{figure}
Addressing this research gap, we develop a flying robot with a unique perching and tilting mechanism shown in \reffig{fig:arche}, capable of exerting large interaction forces while maintaining precise and accurate tool positioning.
Actuation complexity and weight reduce significantly by enabling the robot to 
 attach itself to vertical walls and reuse rotor thrust for tilting and tool feed.
The power tool tightly integrates into the drone frame in a compact design. 
A linear tool table compensates for imprecise perching due to aerodynamic disturbances close to walls.

In summary, this paper presents an aerial platform for construction applications and other force-intense, invasive tasks. The system and the underlying unique tilting process are developed and evaluated. Thorough experiments show accurate pinpointing of target positions, precise manipulation, and versatile tool application. Further, we discuss design limitations and possible improvements. 

%%%%%%%%%%%%%%%%%%%%%%%%%%%%%%%%%%%%%%%%%%%%%%%%%%%%%%%%%%%%%%%%%%%%%%%%%%%%%%%%%%%%%%%%%%%%%%%%%%%%%%%%%%%%%%%%%%%%%%%%
\section{Related Work}
\label{sec:related_work}
\begin{table*}
\vspace{3mm}
\centering
\begin{tabulary}{0.9\textwidth}{@{}lRRRRRLL@{}} 
\toprule 
Design & Reaction Force & Repeated Precision (Samples)  & System Diameter & Payload Capacity & Total Weight & Operation Mode & Demonstrated\newline Application \\ 
\midrule 
         Delta arm hexacopter~\cite{tzoumanikas2020aerial} & \SI{0.25}{\newton} & - & \SI{0.6}{\metre} & - & \SI{2.6}{\kilo\gram} & In-flight & Layouting \\
         Tool tip hexacopter~\cite{zhang2022implementation} & - & \SI{180}{\milli\metre} (N=6) & \SI[round-mode=places,round-precision=1]{0.74}{\metre} & - & \SI{1.6}{\kilo\gram} & In-flight & \ac{NDT} \\
         \acs*{VT} quadcopter~\cite{ding2021design} & \SI{2}{\newton} & \SI[round-mode=places,round-precision=0]{2.5}{\milli\metre} (N=3) & - & - & - & In-flight & drilling \\
         Tool tip quadcopter~\cite{stephens2022integrated} & \SI{4}{\newton} & - & \SI[round-mode=places,round-precision=1]{0.75}{\metre} & \SI{0.15}{\kilo\gram} & \SI{1.9}{\kilo\gram} & In-flight & Sensors \\
         Tool tip hexacopter~\cite{meng2019hybrid} & \SI{5}{\newton} & - & \SI{0.6}{\metre} & - & \SI{2.4}{\kilo\gram} & In-flight & \ac{NDT} \\
         \acs*{MDT} hexacopter~\cite{ryll20176d,tognon2019truly} & \SI{5}{\newton} & - & \SI[round-mode=places,round-precision=1]{1.05}{\metre} & - & \SI{1.8}{\kilo\gram} & In-flight & \ac{NDT} \\
         \acs*{VT} hexacopter~\cite{pantic2019milling,bodie2020active} & \SI{20}{\newton} & - & \SI[round-mode=places,round-precision=1]{0.83}{\metre} & \SI[round-mode=places,round-precision=1]{0.15}{\kilo\gram} & \SI[round-mode=figures,round-precision=2]{5}{\kilo\gram} & In-flight & \ac{NDT}, milling \\
         \acs*{MDT} octocopter~\cite{trujillo2019novel} & \SI{20}{\newton} & - & \SI[round-mode=places,round-precision=1]{2}{\metre} & - & \SI{25}{\kilo\gram} & In-flight & \ac{NDT} \\
         \acs*{VT} tricopter~\cite{watson2021dry} & \SI{30}{\newton} & \SI{36}{\milli\metre} (N=11) & \SI[round-mode=figures,round-precision=2]{1.3}{\metre} & \SI{1}{\kilo\gram} & \SI[round-mode=places,round-precision=1]{5}{\kilo\gram} & In-flight & \ac{NDT} \\
         \acs*{MDT} quadcopter~\cite{watson2022techniques} & \SI{8}{\newton} & \SI[round-mode=places,round-precision=0]{7.5}{\milli\metre} (N=2) & \SI{0.4}{\metre} & \SI{0.3}{\kilo\gram} & \SI{2.1}{\kilo\gram} & Crawling & \ac{NDT} \\
         Roller-arm quadcopter~\cite{meng2019hybrid} & \SI{5}{\newton} & - & \SI{0.6}{\metre} & - & \SI[round-mode=figures,round-precision=2]{2.367}{\kilo\gram} & Crawling & \ac{NDT} \\
         %Perching quadcopter~\cite{wopereis2016mechanism} & - & - & \SI[round-mode=places,round-precision=1]{0.99}{\metre} & - & \SI{1.8}{\kilo\gram} & Perching & None \\
         Octocopter~\cite{sun2021switchable} & \SI{74}{\newton} & - & \SI{1.6}{\metre} & \SI{2}{\kilo\gram} & \SI[round-mode=figures,round-precision=2]{18.6}{\kilo\gram} & Perching & Cleaning \\
         Tilting quadrotor (proposed) & \SI{150}{\newton} & \SI[round-mode=places,round-precision=0]{5.68}{\milli\metre} (N=9) & \SI[round-mode=places,round-precision=1]{1}{\metre} & \SI[round-mode=places,round-precision=1]{2.3}{\kilo\gram} & \SI{13}{\kilo\gram} & Perching & Power tools \\
\bottomrule
\end{tabulary}
\vspace{2px}
\caption{Drone designs that demonstrated contact-based work on vertical surfaces, sorted by operation mode.}
\label{tab:systemcomparison}
\end{table*}
% Drones
Thus far, heavy-duty manipulation with drones has only been shown for drilling holes into the ground~\cite{sun2018unmanned}.
Here, the platform weight provides the feed force.
Work on walls with sustained contact is an active research field.
Applying sufficient horizontal forces while overcoming gravity brings a flying platform to its physical limits.

Most of the manipulation research focuses on performing work \textit{in-flight}.
Common designs extend drones with a single thrust direction, e.g., tri-, quad- or hexacopter, with end effectors.
Because of translational and rotational motion coupling, applying precise, continuous contact force with these devices is challenging.
To alleviate the contact, researchers have been implementing dexterous arms~\cite{kim2015operating,orsag2017dexterous}, compliant magnetic adhesion~\cite{mattar2018development}, laser-guiding~\cite{gonzalez2020uav,zhang2022implementation}, support structures~\cite{albers2010semi,gonzalez2020uav}, delta arms~\cite{tzoumanikas2020aerial} or compliant tool tips~\cite{stephens2022integrated,zhang2022implementation}.
However, these apparatuses must be mounted with sufficient offset from the rotors to avoid collision and wall effects. 
The imbalance only allows small payloads, e.g., for drawing or \acf{NDT}.

\Ac{MDT} or \ac{VT} drones, which can independently control attitude, can have the payload closer to the center of mass, stay in continuous contact more easily and apply more force~\cite{ryll20176d,tognon2019truly,bodie2020active,trujillo2019novel,watson2022techniques}.
However, the redundancy in control comes at the price of increased mechanical complexity and platform weight, effectively reducing payload capacity and available reaction force. Therefore, these platforms do not provide the force to operate power tools interacting with hard materials. Indeed, the \ac{VT} hexacopter presented in~\cite{pantic2019milling} and the \ac{VT} quadcopter in~\cite{ding2021design} showcase milling and drilling only in polystyrene and plywood.

% Crawling 
\textit{Crawling} approaches can significantly increase forces and tool placement precision.
Crawling drones intentionally use contact to stabilize their dynamics by utilizing the rotor thrust to defy gravitational forces, press against the surface, and move~\cite{wopereis2018multimodal,meng2019hybrid,jiang2020real,lanegger2022aerial,watson2022techniques}.
However, on vertical walls, a significant part of the total thrust is required to stay airborne, leaving an insufficient margin to perform power tool work.

% Perching
Ultimately, \textit{perching} is the only way to exert significant forces onto walls.
Perching also has the advantage of completely decoupling flight control and tool operation. 
The decoupling extends operation time significantly. 
Moreover, state estimation, which might deteriorate under strong contact vibrations, is irrelevant during perching.
Aerial perching methods are form-closure, negative fluid pressure, penetration, and dry, magnetic, or electrostatic adhesion~\cite{meng2022aerial}.
Only form-closure~\cite{ruiz2022sophie}, magnetic adhesion~\cite{mattar2018development}, and negative fluid pressure~\cite{wopereis2016mechanism,sun2021switchable} have shown resistance to strong reaction forces.

Given the flat and smooth concrete walls encountered in construction environments, suction is the most efficient and flexible approach.
In particular, bellow suction cups, combined with a membrane pump and electromagnetic valve, are a lightweight solution robust to imprecise approach angles~\cite{liu2020adaptive}.
Related research has already investigated suction cups for high lateral load applications, such as drilling~\cite{kessens2019toward}, but a research gap exists in actual power tool integration and positioning.
In fact, the work in~\cite{sun2021switchable} is one of the first to use suction for performing work on vertical surfaces with an aerial vehicle. The system consists of a regular octocopter with an attached window-cleaning device. 
However, this design has no mechanism to position the end effector parallel to the wall and requires continuous thrust during perching.

\reftab{tab:systemcomparison} summarizes the discussed existing systems that demonstrated work on vertical walls.
Here, \textit{reaction force} is the continuous force applied on a vertical surface, and \textit{repeated precision} corresponds to the maximum deviation from the mean error of consecutive tool applications.
%The table highlights that only our proposed design can generate \SI{150}{\newton} continuous force and carry heavy tools while remaining compact and precise. 
The table highlights that our proposed design exceeds all other approaches in terms of generated force (up to \SI{150}{\newton}) and payload capacity (up to \SI{2.3}{\kilo\gram}), while remaining compact (\SI{1}{\metre}) and precise (\SI[round-mode=places,round-precision=0]{5.68}{\milli\metre}).
The table also shows that most research only considers some of the four metrics. In particular, precision and payload capacity are often not evaluated and standardized tests are missing.

%%%%%%%%%%%%%%%%%%%%%%%%%%%%%%%%%%%%%%%%%%%%%%%%%%%%%%%%%%%%%%%%%%%%%%%%%%%%%%%%%%%%%%%%%%%%%%%%%%%%%%%%%%%%%%%%%%%%%%%%
\section{System Design}
\begin{figure}[b]
\center
\includegraphics[width=\linewidth]{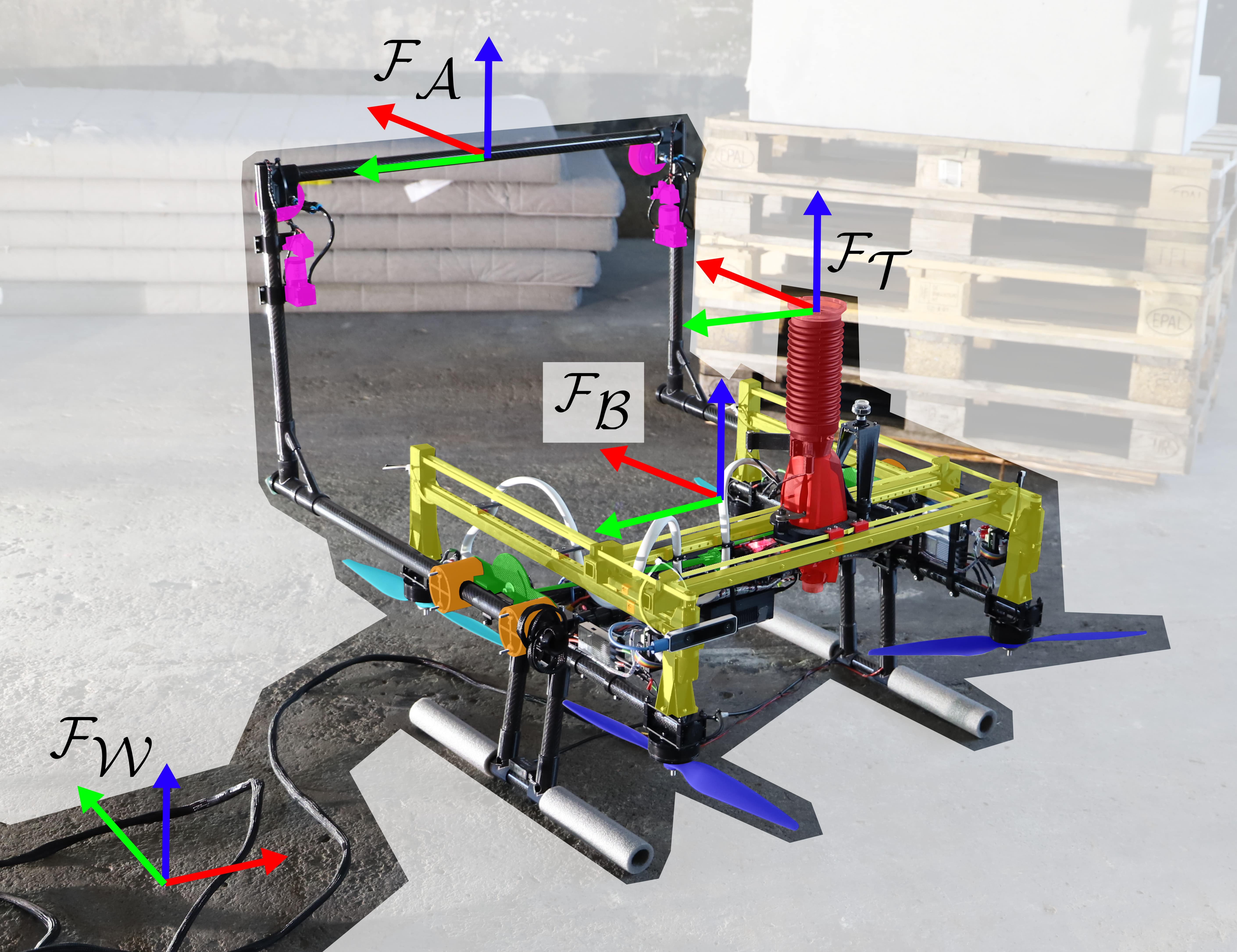}
\caption{The aerial robot with the front and back propellers (light and dark blue), the hammer drill tool (red), the tool positioning system (yellow), the hinge mechanism (green), and its slide bearings (orange), and the suction cups (magenta). The xyz-coordinate frames follow RGB color-convention.}
\label{fig:system_overview}
\end{figure}

Our aerial robot tightly integrates a quadrotor base with a tool positioning and an attachment system. The attachment system is connected to the quadrotor base by a hinge mechanism allowing the base to rotate when perched. In the following, we first introduce the different coordinate frames the system defines and then describe the individual system components in greater detail. The system dimensions and weight is summarized in \reftab{tab:dimensions}, and the full design of the aerial robot with the individual components colored-coded is displayed in \reffig{fig:system_overview}.
\begin{table}[t]
\vspace{3mm}
\begin{center}
\begin{tabular}{ll} 
 \hline
 Quadrotor base weight  & \SI{6.6}{\kilo\gram} \\ 
 Positioning system weight & \SI{2.2}{\kilo\gram} \\ 
 Tool weight & \SI{2.3}{\kilo\gram} \\
 Tether weight & \SI{0.2}{\kilo\gram\per\metre}\\
 % Total flying weight & 11169\,g \\ 
 Height & \SI{0.77}{\metre}\\
 Width & \SI{0.73}{\metre} \\
 Length & \SI{1.22}{\metre} \\
 Propeller diameter & \SI{0.48}{\metre} \\
 \hline
\end{tabular}
\vspace{2px}
\caption{Weight and dimensions of the aerial robot.}
\label{tab:dimensions}
\end{center}
\end{table}

\subsection{Nomenclature}
We introduce coordinate frame $i$ as $\mathcal{F}_i \coloneqq \left\{\mathbf{O}_i, \mathbf{x}_i, \mathbf{y}_i, \mathbf{z}_i \right\}$, with origin $\mathbf{O}_i$, and primary axes $\mathbf{x}_i, \mathbf{y}_i, \mathbf{z}_i$. The proposed system has four coordinate frames. The body-fixed frame $\mathcal{F}_\mathcal{B}$ whose origin $\mathbf{O}_\mathcal{B}$ is attached to the robot's \ac{COM} and whose axes are oriented according to the \ac{FLU} convention. The pose of the aerial robot is given with respect to a fixed world-frame $\mathcal{F}_\mathcal{W}$, with arbitrary origin $\mathbf{O}_\mathcal{W}$ and $\mathbf{z}_\mathcal{W}$ aligned with the negative gravity direction. The attachment frame $\mathcal{F}_\mathcal{A}$ is centered between the two suction cups, with $\mathbf{x}_\mathcal{A}$ pointing in the direction of the suction force and $\mathbf{y}_\mathcal{A}$ pointing from the right to the left suction cup. Finally, the tool frame $\mathcal{F}_\mathcal{T}$ is attached to the tool with $\mathbf{O}_\mathcal{T}$ at the tooltip, and its axes are aligned with $\mathcal{F}_\mathcal{B}$.
The frames can move relative to each other, as stated in the following descriptions and summarized in \reftab{tab:states}.

\subsection{Quadrotor Base}
The base of the aerial robot is an H-shaped quadrotor configuration. The widespread use of this configuration allows simple and well-understood control for flight. Furthermore, the H-shape provides a large rectangular area serving as a workspace. The shape allowed us to tightly integrate the tool positioning and attachment system into the airframe, stiffening the frame and saving weight. A custom-built power tether is connected to the flying platform to provide continuous power during all operations. During the flight, a cascaded PD loop tracks the operator control reference using the attitude, linear velocity, and heading provided by a \ac{VIO} sensor. Additional switches on the \ac{RC} change between different control modes depending on the mode of operation (explained in more detail in \refsec{sec:operation_modes}). A camera stream attached to the power tool provides feedback for precise tool positioning. The output of the camera stream is shown in \reffig{fig:precision_sensors}. The deployed software runs onboard on an \textit{UP Xtreme i11}. All components onboard the aerial robot are listed in \reftab{tab:components}.
\begin{table}[bt]
\vspace{3mm}
\begin{center}
\begin{tabular}{ l l } 
 \hline
 Computer & UP Xtreme i11 - i7 processor\\
 Propulsion Motor & Maxon EC 69 flat UAV \\ 
 ESC's & Maxon UAV-ESC 52/30 \\ 
 Propellers & APC 19x8E \\ 
 State Estimation & Intel Realsense T265 \\ 
 Servo Motor & Robotis Dynamixel XM430-W350-R \\ 
 Suction Cups & Piab B75XP \\
 Membrane Pumps & KNF NMP 830 HP \\
 \hline
\end{tabular}
\vspace{2px}
\caption{Components of the aerial robot.}
\label{tab:components}
\end{center}
\end{table}

\subsection{Attachment System}
The attachment system is the interface between the wall and the flying platform. It consists of an L-shaped carbon rod structure with two suction cups, each actuated by a lightweight membrane pump. For rapid detaching, two actuated valves provide orifices. The valves and pumps are mounted directly below the suction cups to keep the hose length minimal and simplify its routing. The attachment system is connected to the quadrotor base by the hinge mechanism.  

\subsection{Hinge Mechanism}
\begin{figure*}%
\vspace{3mm}
\centering
\subfigure[Flight and locked hinge.]{%
\label{fig:phases_col_a}%
\includegraphics[height=2.35in]{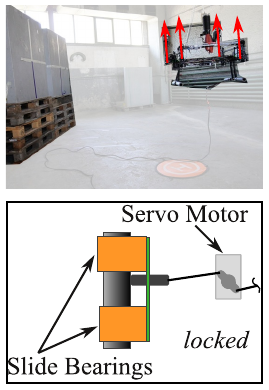}}%
\quad
\subfigure[Perching and locked hinge.]{%
\label{fig:phases_col_b}%
\includegraphics[height=2.35in]{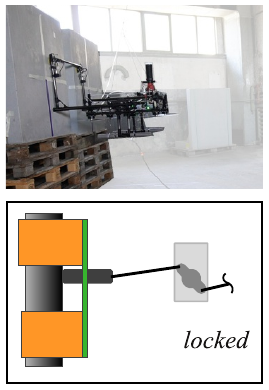}}%
\quad
\subfigure[Rotation and released hinge.]{%
\label{fig:phases_col_c}%
\includegraphics[height=2.35in]{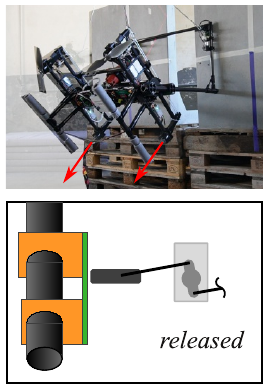}}%
\quad
\subfigure[Manipulation and rotation-locked hinge.]{%
\label{fig:phases_col_d}%
\includegraphics[height=2.35in]{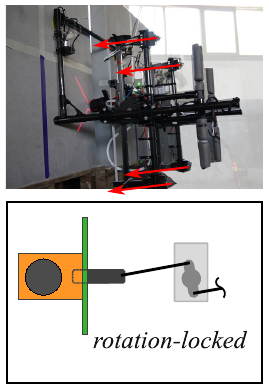}}%
\caption{A typical operation with the corresponding rotor thrust vectors (top row). The bottom row displays a schematic of the locking mechanism's state viewed along $-\mathbf{z}_{\mathcal{B}}$, with locking pin and servo attached to $\mathcal{F}_{\mathcal{B}}$, and carbon plate (green) and slide bearings (orange) attached to $\mathcal{F}_{\mathcal{A}}$ (see \reffig{fig:system_overview}).}
\label{fig:flight_phases}
\end{figure*}

The hinge mechanism connects the attachment structure to the quadrotor. Each side of the hinge mechanism consists of one rotational and two sliding bearings providing the aerial robot with two additional degrees of freedom: rotation around $\mathbf{y}_\mathcal{B}$ defined as $ \theta_{\mathcal{A}\mathcal{B}}^y\in \left[0, 90^\circ\right]$ and translation along $\mathbf{x}_\mathcal{A}$ given by $p_{\mathcal{A}\mathcal{B}}^x \in \mathbb{R}$. A single servo motor with push-rods transitions the hinges on both sides between three states; locked, released, and rotation-locked, also outlined in \reffig{fig:flight_phases}. When the rods are fully extended, the locking pin fully constrains both degrees of freedom through a form-locking connection with the carbon plate ($\theta_{\mathcal{A}\mathcal{B}}^{y})$ and friction contact with the slide tubes ($p_{\mathcal{A}\mathcal{B}}^x$).
In the released state, the pins are fully released, and both degrees of freedom are unlocked. The pins are partially extended to transition into the rotation-locked state, which allows linear but blocks rotational motion. In this state, the pins only form a form-locking connection with the hinge without contacting the carbon tubes. 
%\add{This closure contact is created with the carbon plate (green in \reffig{fig:system_overview} and \reffig{fig:flight_phases} of the hinge, which is attached to the slide bearings and has holes through which the pins can prevent rotational motion and additionally linear motion .}

\subsection{Tool Positioning System}
The tool positioning system is a belt-driven gantry system rigidly attached to the quadrotor and powered by two servo motors. It allows the tool to be moved freely in the xy-plane of $\mathcal{F}_{B}$ defined as $\boldsymbol{p}_{\mathcal{B}\mathcal{T}}^{xy} \in \mathbb{R}^2$. As such, it allows the operator to position the tool precisely at the desired location before it interacts with the surface. Two lasers and a camera are additionally mounted on the tool to project the tooltip position onto the wall and stream the laser cross to the operator. Note that during operation the tool position changes to adjust the \ac{COM} for flying and rotating.

\section{Operation Modes}
\label{sec:operation_modes}
\begin{table}[t]
    \centering
    \begin{tblr}{l|X[c]|X[c]|X[c]}
         & $\theta_{\mathcal{A}\mathcal{B}}^y$ & $p_{\mathcal{A}\mathcal{B}}^{x}$ & $\boldsymbol{p}_{\mathcal{B}\mathcal{T}}^{xy}$ \\
         \hline
         Flight               & Locked & Locked & Locked \\ 
          Perching            & Locked & Locked & Locked \\ 
         Rotation             & Free   & Free   & Locked \\
         Surface Manipulation & Locked & Free   & Free \\
    \end{tblr}
    \vspace{2px}
    \caption{State of the individual degrees of freedom at different operation modes. }
    \label{tab:states}
\end{table}
%%%%%%%%%%
% PROCESS DESCRIPTION
One operator controls the whole aerial robot using a single \ac{RC} during a mission. 
In general, it is possible to split an individual mission into five separate phases; see also \reffig{fig:flight_phases}:

\textit{Flight}: The operator commands a heading and velocity reference to the flight controller in order to navigate the robot into proximity of the target location.  Additionally, an onboard \ac{VIO} sensor provides the controller with the required current odometry of the robot. 

\textit{Perching}: In proximity to the wall, the operator turns on the vacuum pumps and approaches the desired target. When the suction cups contact the wall the robot perches onto the target. The operator then  slowly ramps down the propellers as now the attachment system holds the robot's full weight.

\textit{Rotation}: The operator enables the rotation mode on the \ac{RC}. The flight controller turns off, and instead, an open-loop pitch control starts. The locking pins are fully released, and the tool moves to the workspace center to facilitate rotation by shifting the \ac{COM}. Commanded thrust is now realized by the front propellers, causing the tool table to rotate. When $\theta_{\mathcal{A}\mathcal{B}}^y = 90^\circ$, the operator engages the locking mechanism, partly locking $\theta_{\mathcal{A}\mathcal{B}}^y$, but still allowing sliding in $p_{\mathcal{A}\mathcal{B}}^{x}$.
Note, the hinge does not slide during this phase because the front propellers only generate thrust downward and away from the wall, i.e., in negative $\mathbf{x}_\mathcal{A}$- and  $\mathbf{z}_\mathcal{A}$-direction.

\textit{Manipulation}: The operator can now use the tool positioning system to move the power tool in $\boldsymbol{p}_{\mathcal{B}\mathcal{T}}^{xy}$ and place it in the desired position. Afterward, the operator evenly spins up all four propellers and turns the tool on. Consequently, the power tool is pushed towards the wall and performs the task. 

\begin{figure*}
\centering
\subfigure[The thrust (black) ramps up until $t = 42$\,s, the pumps disable at $t = 49$\,s, and the valves open at $t = 54$\,s. RGB show the position of the body with respect to the wall $\boldsymbol{p}_{\mathcal{W}\mathcal{B}}$, with $\mathbf{x}_\mathcal{W}$ pointing away from the wall.]{%
\label{fig:undocking_plot}%
\includegraphics[width=0.45\textwidth]{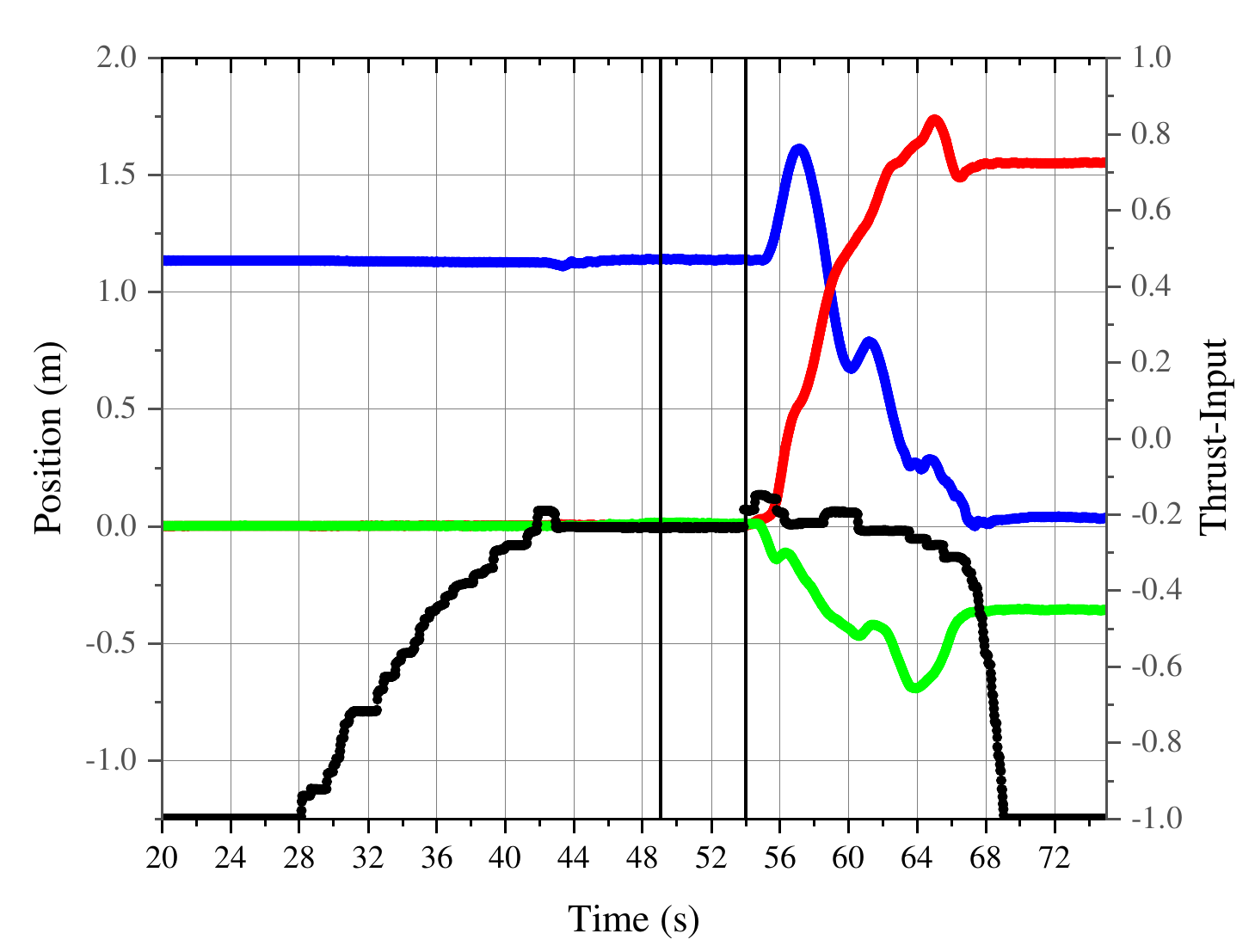}}%
\quad
\subfigure[The robot before detaching from the wall, with coordinate frame used for \reffig{fig:undocking_plot}]{%
\label{fig:undocking_coordframe}%
\includegraphics[width=0.45\textwidth]{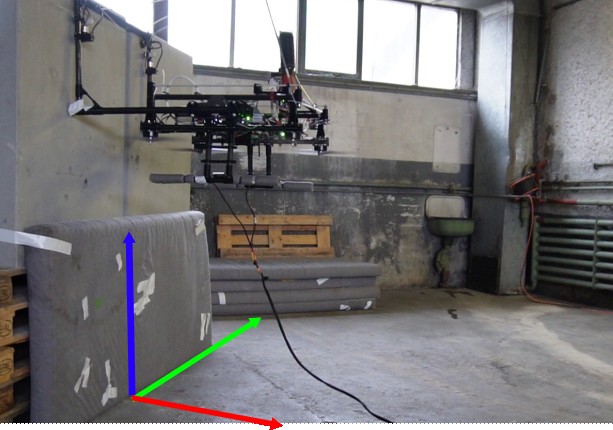}}%
\caption{Detachment from the wall and landing.}
\label{fig:undocking}

\end{figure*}

\textit{Detachment}:
After the tool application, the operator executes all previous steps in reverse. The tool retracts by spinning all propellers backward, the locking pins are fully released, and the robot rotates back to $\theta_{\mathcal{A}\mathcal{B}}^y = 0^\circ$ by spinning the front propellers. 
Afterward, the hinge mechanism is locked, and the tool is moved to the quadrotor's back to center the \ac{COM} for flight. Before take-off, the state-estimator, provided by the \ac{VIO} sensor, and the in-flight controller are re-started. 
The operator spins up the rotors to hover thrust and commands a velocity in negative-$\mathbf{x}_\mathcal{B}$ direction such that the robot leans away from the wall. Finally, \reffig{fig:undocking} shows that the aerial robot detaches and moves away from the wall when the pressure valves open. 

\section{Experimental Validation}
In the following, we evaluate the main design criteria of this aerial robot: high reaction forces, compact design, and accurate and precise tool application.
The experiments were performed indoors on a CreaBeton M8110 119420 smooth concrete slab, also shown in \reffig{fig:flight_phases}.
The indoor application and safety tether reduced risk and regulation constraints.
Furthermore, the system was validated outdoors on smooth concrete walls and wood in a training facility shown in \reffig{fig:arche}. 

%The concrete slab is sandblasted and contains pores with a diameter of up to \SI{15}{\milli\metre} and a depth of \SI{5}{\milli\metre}. Per the manufacturer's specifications, there are no more than five pores with a diameter of \SIrange{10}{15}{\milli\metre} on an area of \SI{100}{\centi\metre\squared}. 

\begin{figure}
\centerline{\includegraphics[width=0.45\textwidth]{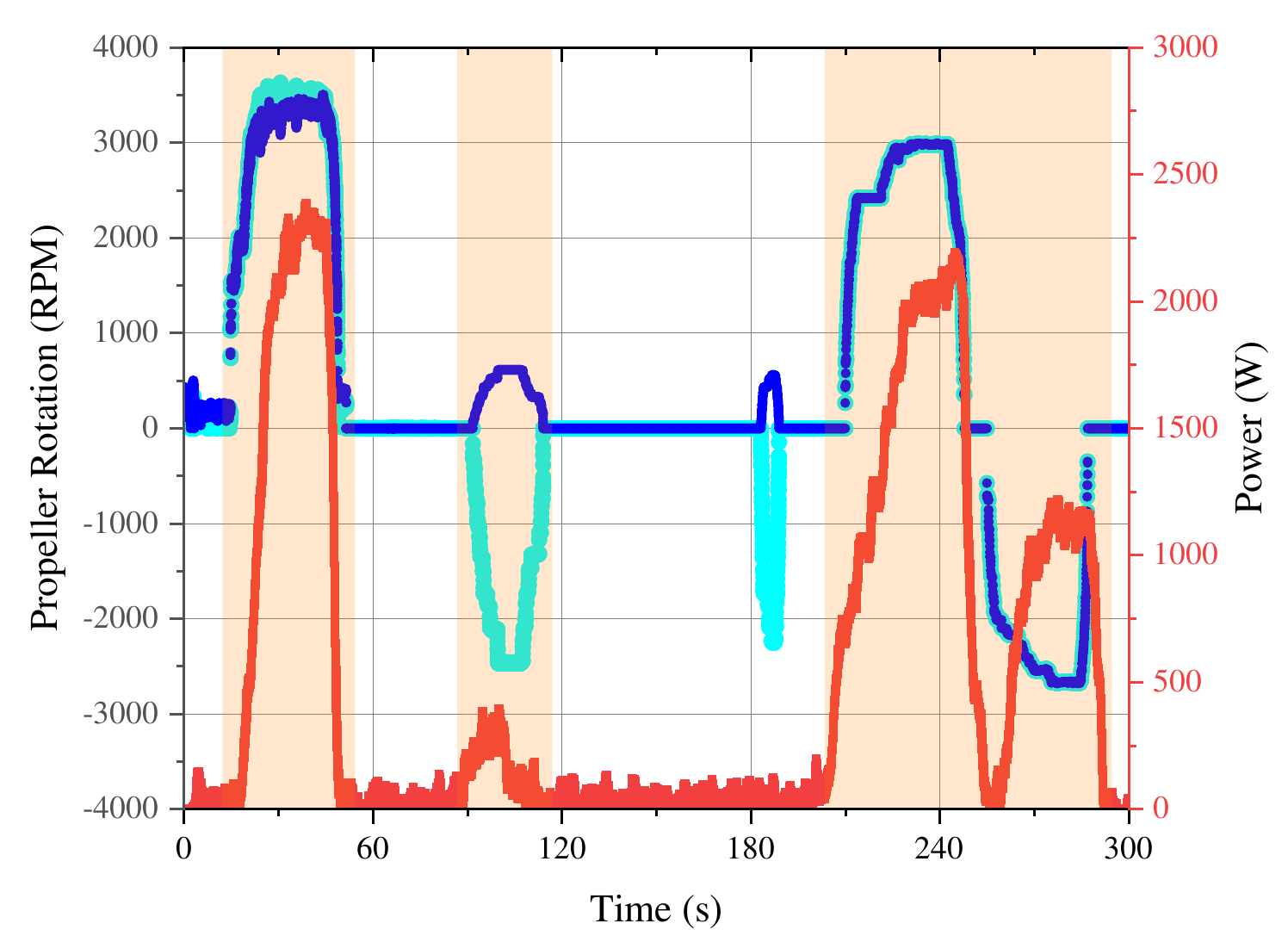}}
\caption{Power consumption (red) and propeller rotation of front (cyan) and back (blue) propeller pair during the main part of a typical drilling operation. The orange sectors are, chronologically: flight, rotation and manipulation including tool movement along $\mathbf{x}_\mathcal{A}$.}
\label{fig:poweromega}
\end{figure}

\subsection{Reaction Forces and Compact, Power-Efficient Design}
The propulsion and power consumption during operation are summarized in \reffig{fig:poweromega}.
During the flight, both propeller pairs have approximately the same \ac{RPM} due to a centered \ac{COM} induced by the power tool positioning in the quadrotor's back, counteracting the attachment system weight.
Once perched after $t = 50$\,$s$, the platform consumes neglectable power.
For rotation, the front propellers spin in a negative direction.
Later, the four propellers rotate at identical speeds during linear tool feed and retraction. The \SI{3000}{\ac{RPM}} during drilling corresponds to roughly \SI{110}{\newton} reaction force, neglecting friction losses.
We measured \SI[separate-uncertainty,multi-part-units=repeat]{150\pm5}{\newton} reaction force at maximum \ac{RPM} in five separate experiments.
The power consumption for flying and manipulation is almost identical at around \SI{2}{\kilo\watt}, showcasing effective motion separation.
An \textit{in-flight} or \textit{crawling} aerial robot design would inevitably be much larger in order to provide the same reaction forces while simultaneously maintaining thrust for gravity compensation. 

Using a power tether capable of continuously delivering \SI{2}{\kilo\watt}, the proposed prototype exhibits potentially unlimited operation time. To prevent mobility impediment and reduce weight at high altitudes, however, the external power supply could be replaced with an on-board \ac{LiPo}. Depending on the specific energy density, a \SI{1}{\kilo\gram} battery would approximately allow \SI{250}{\second} of non-stop flight/manipulation time.

\subsection{Robust Tool Positioning}
\begin{figure*}[b]
\centering
\subfigure[Perching on target cross.]{%
\label{fig:perching_overview}%
\includegraphics[height=2.8in]{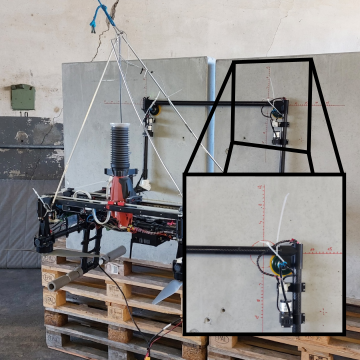}}%
\quad
\hspace{15mm}
\subfigure[Perching points (black) and reachable tool workspace (blue). The black crosses indicate the target perching points, the red plus is the desired manipulation location.]{%
\label{fig:perching_results}%
\includegraphics[height=2.9in]{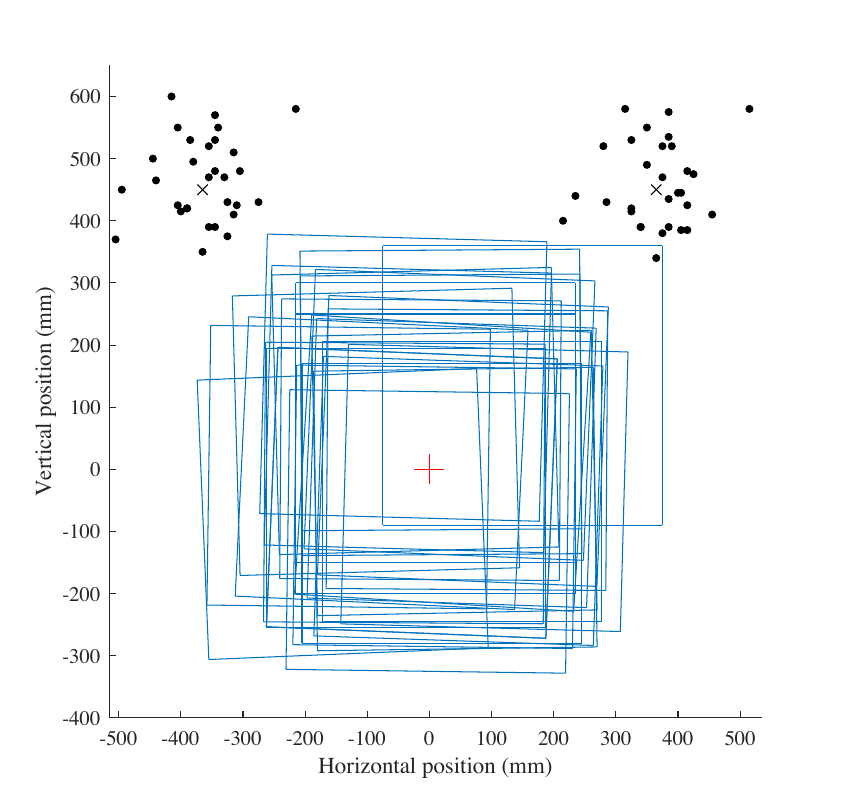}}%
%\centerline{\includegraphics[width=0.45\textwidth]{perching_v4.eps}}
\caption{Perching locations for $30$ consecutive perching attempts.}
\label{fig:perching_accuracy}
\end{figure*}

Aerodynamic disturbances close to walls reduce the perching precision.
However, the tool table compensates for flight inaccuracies.
To show this, we demonstrate $30$ consecutive wall attachments.
The operator visually steered the quadrotor to attach the suction cups to two target crosses shown in \reffig{fig:perching_overview}.
The suction cup center locations were manually measured with respect to the target crosses.
\reffig{fig:perching_results} shows the docking precision deviated around \SI{\pm10}{\centi\metre} in any direction.
Nevertheless, our tool table can position the tool at any point within a \SI[parse-numbers=false]{150 \times 210}{\milli\metre\squared} window, including the desired tool position at the nominal tool table center. 

We believe docking and, thus, the effective tool workspace can be improved in the future. 
Carbon propellers will provide better control through a higher \ac{F/W}, and automated flight navigation will eliminate human RC errors and viewpoint constraints.

\subsection{Tool Application Accuracy and Precision}
\begin{figure*}%
\vspace{3mm}
\centering
\subfigure[Photo of lasercross, hole pattern, and fully retracted drill with a dust collector. Note the offset between lasercross and hole.]{%
\label{fig:precision_overview}%
\includegraphics[height=1.33in]{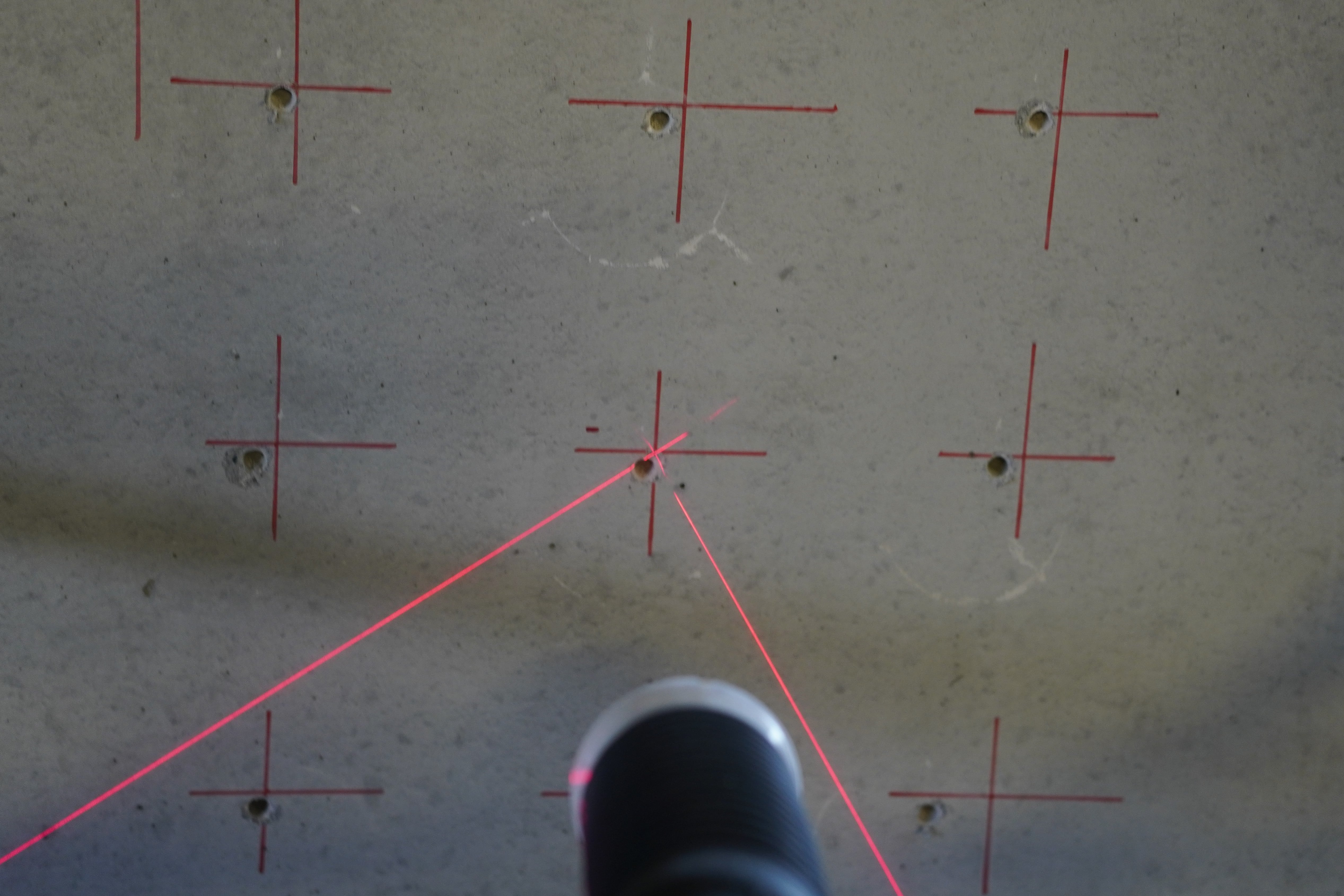}}%
\quad
\subfigure[The camera feed that the operator uses to position the tool. One pixel relates to \SI{4}{\milli\metre} on the concrete wall.]{%
\label{fig:precision_operator_view}%
\includegraphics[height=1.33in]{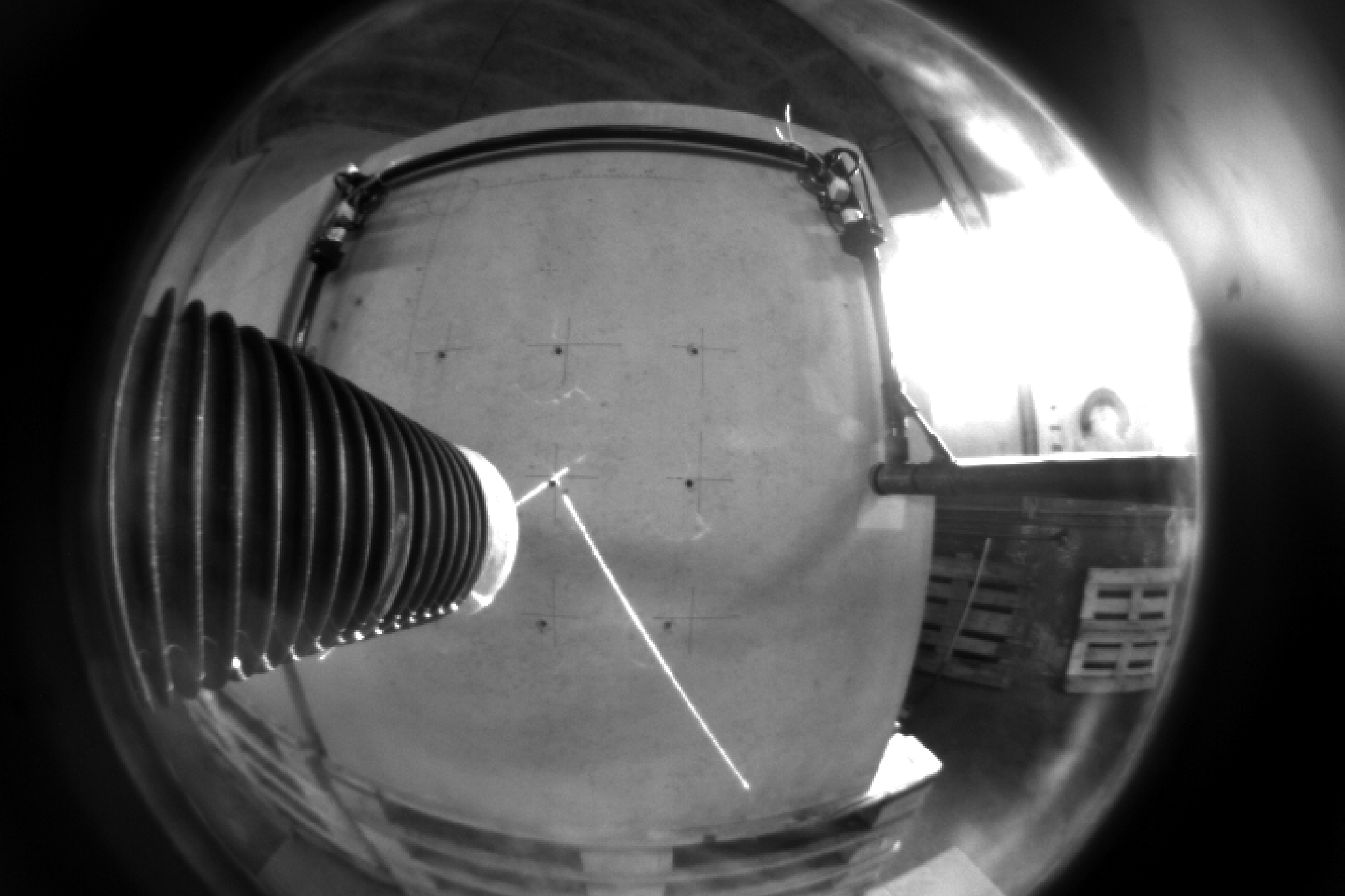}}%
\quad
\subfigure[Hole offset (grey) relative to the target location with one outlier (blue), and inlier maximum offset (circle) around the mean offset (red).]{%
\label{fig:drilling_precision}%
\includegraphics[height=1.395in]{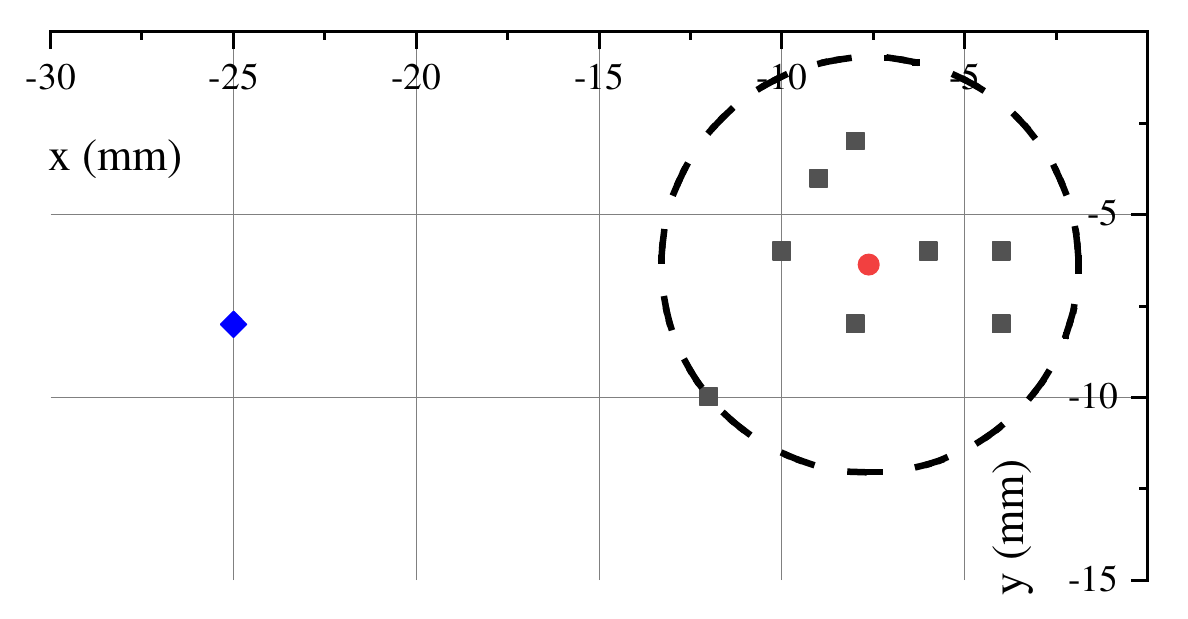}}%
\caption{Linear tool feed accuracy and precision evaluation via drilling a \si{3\times3} hole pattern.}
\label{fig:precision_sensors}
\end{figure*}

We show the accuracy and precision of the tool application by drilling the \si{3\times3} hole pattern shown in \reffig{fig:precision_overview}.
The drilling application is a realistic and challenging use case of the proposed system.
\reffig{fig:precision_operator_view} shows the operator's view of aligning the gantry system with the target drilling location.
After alignment, constant thrust drives the drill into the wall.
The operator does not re-adjust the drill position while feeding forward, which ensures evaluation of the entire tool positioning mechanism, including sliding.
\reffig{fig:drilling_precision} shows the manually measured holes relative to their target locations.
The holes' accuracy, i.e., the average offset from the target, is \SI[round-mode=places,round-precision=0]{9.94}{\milli\metre}.
The precision, i.e., the maximum offset from the mean, is \SI[round-mode=places,round-precision=0]{5.68}{\milli\metre}.
Note that we excluded the blue outlier that originated due to suction cup slipping induced by the low temperatures in the last experiment.

The systematic offset towards the bottom left comes from misalignment between the cross-section of the two line lasers and the tooltip.
The imprecision stems from sliding backlash, thrust imbalance, and low camera resolution.
The accuracy could improve with a better camera, line-laser, and tool calibration.
Actively controlling the tool feed, e.g., by visual servoing~\cite{ding2021design}, would drastically improve the precision. 
Still, the high precision shows that the proposed linear guidance is a practical mechanical solution, and simply ramping up the thrust delivers reliable results.
The experiment also highlights the repeated tool application over the entire workspace. 

\begin{comment}
\begin{figure}
\centerline{\includegraphics[width=0.45\textwidth]{undocking_coordframe_V1.jpg}}
\caption{The coordinate frame with RGB-convention used for the plot in fig. \ref{fig:undocking}.}
\label{fig:undocking_coordframe}
\end{figure}    
\end{comment}

% \subsection{Versatility}
% Adapting different tools is simple, and the operation can be performed analogously. To demonstrate the vast capabilities of the system, an impact wrench was mounted on the tool positioning system. In \reffig{fig:screwing} we demonstrate screw placement in the previously drilled holes. The screw is attached to the impact wrench with a small piece of double-sided tape to prevent the screw from falling out. Using the linear table and the line-lasers, the screw can be aligned with a previously drilled hole and the feed force as well as screwing action is performed analogously to the drilling operation. 

\subsection{Discussion}
\begin{figure}
\centerline{\includegraphics[width=0.45\textwidth]{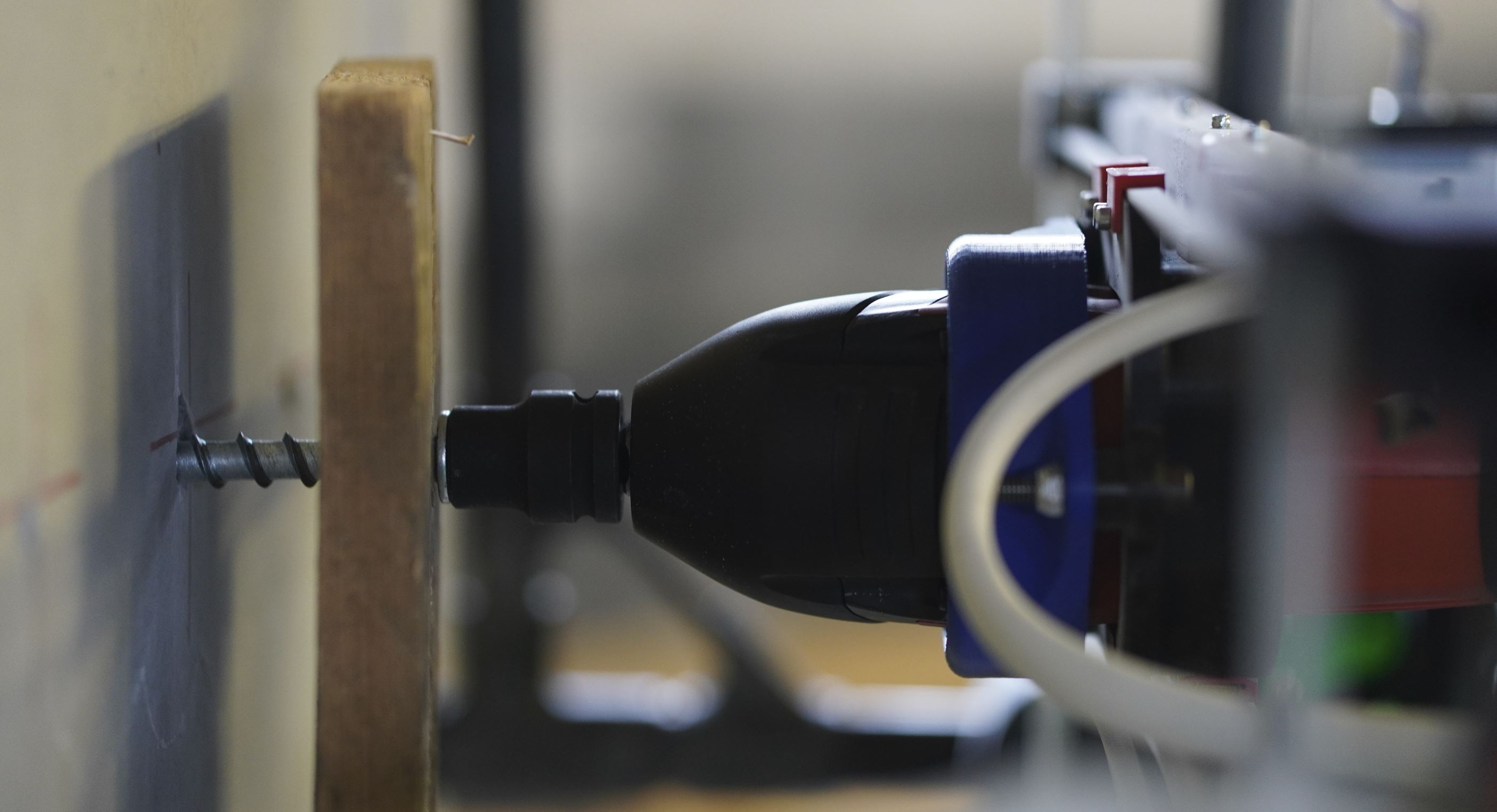}}
\caption{Screw placement with an impact wrench into a previously drilled hole for mounting a wooden board to the concrete wall.}
\label{fig:screwing}
\end{figure}

The results stated in the previous section confirm the prototypes suitability for performing demanding power tool work on construction sites.
Accuracy and precision are particularly important, as multistage workflows often demand repeated application of different tools at the same locations.
To demonstrate such versatility of the system, an impact wrench was mounted on the tool positioning system for screw placement in a previously drilled hole (see \reffig{fig:screwing}).

Despite the experimentally illustrated capabilities of the developed prototype, certain considerations must be made for future developments.
Most importantly, the suction cups tend to slip on dust-covered walls, as well as in the presence of large cracks or protrusions.
Although different designs exist to increase compliance towards rough surfaces (e.g. increased area or greater lip), this will affect overall weight, as well as time between first contact and full adhesion. 
%The suction cups do not perform well in all environments. In particular, dust-covered walls can cause one or both suction cups to glide along the surface \add{until a suitable place for adhesion has been found}. \add{Other surface unevenness such as large cracks or protrusions can cause for weaker attachment. The overall surface topology must be sufficiently flat such that the bellows suction cups can adapt to a change in plane orientation between the two suction cups. By choosing the geometry and material of the suction cups carefully, these issues can be partially remedied. This can however incur a weight penalty or prolong the time between first contact and full adhesion onto the wall. Using a bellows suction cup increases compliance towards non-perpendicular attachment orientation and a large lip creates resilience towards surface irregularities.} 
Further, the currently used suction cups are less compliant below \SI{10}{\celsius}, causing similar slippage even on clean surfaces. 

The hinge mechanism is also heavily temperature dependant, as \ac{PLA} components shrink, causing increased steel locking pin friction. %In general, managing friction is no simple task, as blocking linear motion is also done through friction.
The dusty environment causes the slide bearings to deteriorate over time. It leads to more frequent jamming during the linear movement towards the wall and more thrust needed for manipulation, as the friction of the bearings must also be overcome. 

The heavy vibrations caused by the drill hammer must be considered when designing such a system, both in terms of mechanical robustness and sensor disturbance. However, no state estimation is required when perched and the estimator can therefore be reset before detachment. %\add{For increased mobility, battery-powered flight is an alternative to the current power supply via tether. At \SI{8}{\metre} height, the tether adds \SI{1.6}{\kilo\gram} weight to the robot. This weight corresponds to a \SI{200}{\watt\hour} \ac{LiPo}. During nominal flight, the robot consumes approximately \SI{2.2}{\kilo\watt} (see \reffig{fig:poweromega}), leading to \SI{5}{\minute} battery flight duration, which is sufficient for short deployments. Remember, perching requires almost no power.}

\section{Conclusion}
This paper presents a novel aerial robot design for power tool work on vertical walls.
The platform's unique perch and tilt mechanism decouples flight and manipulation. The decoupling enables exerting of high reaction forces and precise tool positioning using a power-efficient, compact design. Experimental evaluation shows reaction forces of up to \SI{150}{\newton}, robust tool positioning under perching uncertainty, and \SI{10}{\milli\metre} drilling accuracy and \SI{6}{\milli\metre} precision. 
We demonstrate the prototype's capability to perform demanding power tool tasks, specifically drilling a hole into concrete and setting an anchor.
However, the design also suits various other vertical surface applications, e.g., tools, sensor placement, \ac{NDT}, or layouting.
Future work will further improve accuracy and precision through tool calibration and visual servoing, enabling precise drill patterns. Mechanical robustification, perching alternatives, and flight control performance will broaden the application area. 

%\addtolength{\textheight}{-4.5cm} % This command serves to balance the column lengths

\begin{comment}
\section*{Acknowledgment}

The preferred spelling of the word ``acknowledgment'' in America is without 
an ``e'' after the ``g''. Avoid the stilted expression ``one of us (R. B. 
G.) thanks $\ldots$''. Instead, try ``R. B. G. thanks$\ldots$''. Put sponsor 
acknowledgments in the unnumbered footnote on the first page. 
\end{comment}

\bibliographystyle{IEEEtran}
\bibliography{literature.bib}

\begin{thebibliography}{10}
\providecommand{\url}[1]{#1}
\csname url@samestyle\endcsname
\providecommand{\newblock}{\relax}
\providecommand{\bibinfo}[2]{#2}
\providecommand{\BIBentrySTDinterwordspacing}{\spaceskip=0pt\relax}
\providecommand{\BIBentryALTinterwordstretchfactor}{4}
\providecommand{\BIBentryALTinterwordspacing}{\spaceskip=\fontdimen2\font plus
\BIBentryALTinterwordstretchfactor\fontdimen3\font minus
  \fontdimen4\font\relax}
\providecommand{\BIBforeignlanguage}[2]{{%
\expandafter\ifx\csname l@#1\endcsname\relax
\typeout{** WARNING: IEEEtran.bst: No hyphenation pattern has been}%
\typeout{** loaded for the language `#1'. Using the pattern for}%
\typeout{** the default language instead.}%
\else
\language=\csname l@#1\endcsname
\fi
#2}}
\providecommand{\BIBdecl}{\relax}
\BIBdecl

\bibitem{bogue2017prospects}
R.~Bogue, ``What are the prospects for robots in the construction industry?''
  \emph{Industrial Robot: An International Journal}, 2017.

\bibitem{gawel2019fully}
A.~Gawel, H.~Blum, J.~Pankert, K.~Kr{\"a}mer, L.~Bartolomei, S.~Ercan,
  F.~Farshidian, M.~Chli, F.~Gramazio, R.~Siegwart \emph{et~al.}, ``A
  fully-integrated sensing and control system for high-accuracy mobile robotic
  building construction,'' in \emph{International conference on intelligent
  robots and systems (IROS)}.\hskip 1em plus 0.5em minus 0.4em\relax IEEE,
  2019, pp. 2300--2307.

\bibitem{shimizu_corporation_2018}
\BIBentryALTinterwordspacing
{Shimizu Corporation}. Robots under autonomous control at robot lab. [Online].
  Available:
  \url{https://www.shimz.co.jp/en/company/about/news-release/2018/2018006.html}
\BIBentrySTDinterwordspacing

\bibitem{din_18202}
{DIN-Normenausschuss Bauwesen (NABau)}, ``Tolerances in building construction -
  buildings,'' DIN Deutsches Institut für Normung e. V., Standard DIN EN
  18202, July 2019.

\bibitem{bouabdallah2004design}
S.~Bouabdallah, P.~Murrieri, and R.~Siegwart, ``Design and control of an indoor
  micro quadrotor,'' in \emph{International conference on robotics and
  automation (ICRA)}, vol.~5.\hskip 1em plus 0.5em minus 0.4em\relax IEEE,
  2004, pp. 4393--4398.

\bibitem{edmonds2020unmanned}
K.~Edmonds and D.~B. Stringer, ``Unmanned vertical take off and landing (vtol)
  propulsion: Scalability of quadcopter rotormotor configurations outside the
  small uas (suas) regime,'' Kent State University Kent United States, Tech.
  Rep., 2020.

\bibitem{kessens2019toward}
C.~C. Kessens, M.~Horowitz, C.~Liu, J.~Dotterweich, M.~Yim, and H.~L. Edge,
  ``Toward lateral aerial grasping \& manipulation using scalable suction,'' in
  \emph{International Conference on Robotics and Automation (ICRA)}.\hskip 1em
  plus 0.5em minus 0.4em\relax IEEE, 2019, pp. 4181--4186.

\bibitem{watson2022techniques}
R.~Watson, T.~Zhao, D.~Zhang, M.~Kamel, C.~MacLeod, G.~Dobie, G.~Bolton,
  A.~Joly, S.~G. Pierce, and J.~Nieto, ``Techniques for contact-based
  structural health monitoring with multirotor unmanned aerial vehicles,''
  \emph{Proceedings of the 13th International Workshop on Structural Health
  Monitoring}, 2022.

\bibitem{pantic2019milling}
M.~Pantic, M.~Brunner, K.~Bodie, R.~Siegwart, and J.~Nieto, ``Floating-base
  milling using omnidirectional micro aerial vehicles,'' \emph{Robotics:
  Science and System XV, Workshop on Aerial Interaction and Manipulation:
  Unsolved Challenges and Perspectives}, 2019.

\bibitem{tzoumanikas2020aerial}
D.~Tzoumanikas, F.~Graule, Q.~Yan, D.~Shah, M.~Popovic, and S.~Leutenegger,
  ``Aerial manipulation using hybrid force and position nmpc applied to aerial
  writing,'' \emph{Proceedings of Robotics Robotics: Science and System XVI},
  2020.

\bibitem{zhang2022implementation}
D.~Zhang, R.~Watson, C.~MacLeod, G.~Dobie, W.~Galbraith, and G.~Pierce,
  ``Implementation and evaluation of an autonomous airborne ultrasound
  inspection system,'' \emph{Nondestructive Testing and Evaluation}, vol.~37,
  no.~1, pp. 1--21, 2022.

\bibitem{ding2021design}
C.~Ding, L.~Lu, C.~Wang, and C.~Ding, ``Design, sensing, and control of a novel
  uav platform for aerial drilling and screwing,'' \emph{IEEE Robotics and
  Automation Letters}, vol.~6, no.~2, pp. 3176--3183, 2021.

\bibitem{stephens2022integrated}
B.~Stephens, H.-N. Nguyen, S.~Hamaza, and M.~Kovac, ``An integrated framework
  for autonomous sensor placement with aerial robots,'' \emph{IEEE/ASME
  Transactions on Mechatronics}, 2022.

\bibitem{meng2019hybrid}
X.~Meng, Y.~He, and J.~Han, ``Hybrid force/motion control and implementation of
  an aerial manipulator towards sustained contact operations,'' in
  \emph{International Conference on Intelligent Robots and Systems
  (IROS)}.\hskip 1em plus 0.5em minus 0.4em\relax IEEE, 2019, pp. 3678--3683.

\bibitem{ryll20176d}
M.~Ryll, G.~Muscio, F.~Pierri, E.~Cataldi, G.~Antonelli, F.~Caccavale, and
  A.~Franchi, ``6d physical interaction with a fully actuated aerial robot,''
  in \emph{International Conference on Robotics and Automation (ICRA)}.\hskip
  1em plus 0.5em minus 0.4em\relax IEEE, 2017, pp. 5190--5195.

\bibitem{tognon2019truly}
M.~Tognon, H.~A.~T. Ch{\'a}vez, E.~Gasparin, Q.~Sabl{\'e}, D.~Bicego,
  A.~Mallet, M.~Lany, G.~Santi, B.~Revaz, J.~Cort{\'e}s \emph{et~al.}, ``A
  truly-redundant aerial manipulator system with application to push-and-slide
  inspection in industrial plants,'' \emph{IEEE Robotics and Automation
  Letters}, vol.~4, no.~2, pp. 1846--1851, 2019.

\bibitem{bodie2020active}
K.~Bodie, M.~Brunner, M.~Pantic, S.~Walser, P.~Pf{\"a}ndler, U.~Angst,
  R.~Siegwart, and J.~Nieto, ``Active interaction force control for
  contact-based inspection with a fully actuated aerial vehicle,'' \emph{IEEE
  Transactions on Robotics}, vol.~37, no.~3, pp. 709--722, 2020.

\bibitem{trujillo2019novel}
M.~{\'A}. Trujillo, J.~R. Mart{\'\i}nez-de Dios, C.~Mart{\'\i}n, A.~Viguria,
  and A.~Ollero, ``Novel aerial manipulator for accurate and robust industrial
  ndt contact inspection: A new tool for the oil and gas inspection industry,''
  \emph{Sensors}, vol.~19, no.~6, p. 1305, 2019.

\bibitem{watson2021dry}
R.~Watson, M.~Kamel, D.~Zhang, G.~Dobie, C.~MacLeod, S.~G. Pierce, and
  J.~Nieto, ``Dry coupled ultrasonic non-destructive evaluation using an
  over-actuated unmanned aerial vehicle,'' \emph{IEEE Transactions on
  Automation Science and Engineering}, vol.~19, no.~4, pp. 2874--2889, 2021.

\bibitem{sun2021switchable}
Y.~Sun, Z.~Jing, P.~Dong, J.~Huang, W.~Chen, and H.~Leung, ``A switchable
  unmanned aerial manipulator system for window-cleaning robot installation,''
  \emph{IEEE Robotics and Automation Letters}, vol.~6, no.~2, pp. 3483--3490,
  2021.

\bibitem{sun2018unmanned}
Y.~Sun, A.~Plowcha, M.~Nail, S.~Elbaum, B.~Terry, and C.~Detweiler, ``Unmanned
  aerial auger for underground sensor installation,'' in \emph{International
  Conference on Intelligent Robots and Systems (IROS)}.\hskip 1em plus 0.5em
  minus 0.4em\relax IEEE, 2018, pp. 1374--1381.

\bibitem{kim2015operating}
S.~Kim, H.~Seo, and H.~J. Kim, ``Operating an unknown drawer using an aerial
  manipulator,'' in \emph{International conference on robotics and automation
  (ICRA)}.\hskip 1em plus 0.5em minus 0.4em\relax IEEE, 2015, pp. 5503--5508.

\bibitem{orsag2017dexterous}
M.~Orsag, C.~Korpela, S.~Bogdan, and P.~Oh, ``Dexterous aerial robots—mobile
  manipulation using unmanned aerial systems,'' \emph{IEEE Transactions on
  Robotics}, vol.~33, no.~6, pp. 1453--1466, 2017.

\bibitem{mattar2018development}
R.~A. Mattar and R.~Kalai, ``Development of a wall-sticking drone for
  non-destructive ultrasonic and corrosion testing,'' \emph{Drones}, vol.~2,
  no.~1, p.~8, 2018.

\bibitem{gonzalez2020uav}
L.~M. Gonz{\'a}lez-deSantos, J.~Mart{\'\i}nez-S{\'a}nchez,
  H.~Gonz{\'a}lez-Jorge, F.~Navarro-Medina, and P.~Arias, ``Uav payload with
  collision mitigation for contact inspection,'' \emph{Automation in
  Construction}, vol. 115, p. 103200, 2020.

\bibitem{albers2010semi}
A.~Albers, S.~Trautmann, T.~Howard, T.~A. Nguyen, M.~Frietsch, and C.~Sauter,
  ``Semi-autonomous flying robot for physical interaction with environment,''
  in \emph{Conference on robotics, automation and mechatronics}.\hskip 1em plus
  0.5em minus 0.4em\relax IEEE, 2010, pp. 441--446.

\bibitem{wopereis2018multimodal}
H.~W. Wopereis, W.~L. Van De~Ridder, T.~J. Lankhorst, L.~Klooster, E.~M. Bukai,
  D.~Wuthier, G.~Nikolakopoulos, S.~Stramigioli, J.~B. Engelen, and
  M.~Fumagalli, ``Multimodal aerial locomotion: An approach to active tool
  handling,'' \emph{IEEE Robotics \& Automation Magazine}, vol.~25, no.~4, pp.
  57--65, 2018.

\bibitem{jiang2020real}
S.~Jiang and J.~Zhang, ``Real-time crack assessment using deep neural networks
  with wall-climbing unmanned aerial system,'' \emph{Computer-Aided Civil and
  Infrastructure Engineering}, vol.~35, no.~6, pp. 549--564, 2020.

\bibitem{lanegger2022aerial}
C.~Lanegger, M.~Ruggia, M.~Tognon, L.~Ott, and R.~Siegwart, ``Aerial layouting:
  Design and control of a compliant and actuated end-effector for precise
  in-flight marking on ceilings,'' \emph{Proceedings of Robotics Robotics:
  Science and System XVIII}, 2022.

\bibitem{meng2022aerial}
J.~Meng, J.~Buzzatto, Y.~Liu, and M.~Liarokapis, ``On aerial robots with
  grasping and perching capabilities: A comprehensive review,'' \emph{Frontiers
  in Robotics and AI}, p. 405, 2022.

\bibitem{ruiz2022sophie}
F.~Ruiz, B.~C. Arrue, and A.~Ollero, ``Sophie: Soft and flexible aerial vehicle
  for physical interaction with the environment,'' \emph{IEEE Robotics and
  Automation Letters}, vol.~7, no.~4, pp. 11\,086--11\,093, 2022.

\bibitem{wopereis2016mechanism}
H.~W. Wopereis, T.~Van Der~Molen, T.~Post, S.~Stramigioli, and M.~Fumagalli,
  ``Mechanism for perching on smooth surfaces using aerial impacts,'' in
  \emph{International symposium on safety, security, and rescue robotics
  (SSRR)}.\hskip 1em plus 0.5em minus 0.4em\relax IEEE, 2016, pp. 154--159.

\bibitem{liu2020adaptive}
S.~Liu, W.~Dong, Z.~Ma, and X.~Sheng, ``Adaptive aerial grasping and perching
  with dual elasticity combined suction cup,'' \emph{IEEE Robotics and
  Automation Letters}, vol.~5, no.~3, pp. 4766--4773, 2020.

\end{thebibliography}

\end{document}